\UseRawInputEncoding
\documentclass[10pt]{article} 
\pdfoutput=1
\usepackage{simpleConference}
\usepackage[utf8]{inputenc} 
\usepackage[T1]{fontenc}    
\usepackage{hyperref}       
\usepackage{url}            
\usepackage{booktabs}       
\usepackage{amsfonts}       
\usepackage{nicefrac}       
\usepackage{microtype}      
\usepackage{xcolor}         
\usepackage{times}
\usepackage{epsfig}
\usepackage{graphicx}
\usepackage{amsmath}
\usepackage{amssymb}
\usepackage{booktabs}
\usepackage{algorithm}
\usepackage{algorithmic}
\usepackage{makecell}
\usepackage{cuted}
\usepackage{capt-of}
\usepackage{caption}
\usepackage{lipsum}
\usepackage{multirow}
\usepackage{enumitem}

\hypersetup{
colorlinks=true,
linkcolor=black,
citecolor=black
}

\begin{document}

\title{Few-shot 3D Shape Generation}

\author{JingYuan Zhu \\
Tsinghua University, China \\
jy-zhu20@mails.tsinghua.edu.cn\\
\and
Huimin Ma \\
University of Science and Technology Beijing, China \\
mhmpub@ustb.edu.cn  \\
\and
Jiansheng Chen \\
University of Science and Technology Beijing, China \\
jschen@ustb.edu.cn  \\
\and
Jian Yuan  \\
Tsinghua University, China \\
jyuan@tsinghua.edu.cn  \\
}

\maketitle
\thispagestyle{empty}

\renewcommand\twocolumn[1][]{#1}
\maketitle
\begin{center}
    \captionsetup{type=figure}
    \includegraphics[width=1.0\textwidth]{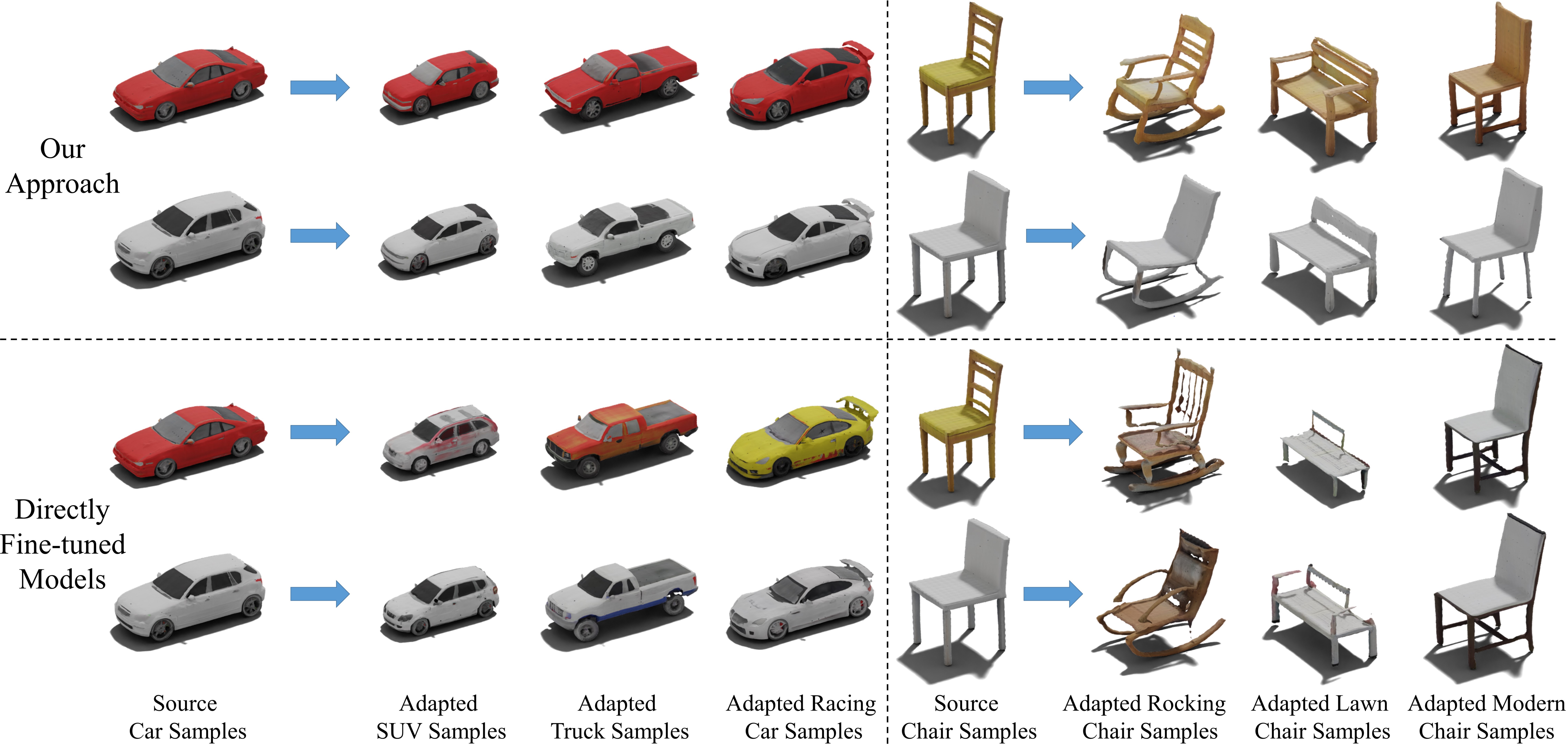}
    \captionof{figure}{Given pre-trained 3D shape generative models, we propose to adapt them to target domains using a few target samples while preserving diverse geometry and texture information learned from source domains. Compared with directly fine-tuned models which tend to replicate the few-shot target samples, our approach only needs the silhouettes of target samples as training data and achieves diverse generated shapes following target geometry distributions but different from target samples.}
    \label{teaser}
\end{center}

\begin{abstract}
 Realistic and diverse 3D shape generation is helpful for a wide variety of applications such as virtual reality, gaming, and animation. Modern generative models, such as GANs and diffusion models, learn from large-scale datasets and generate new samples following similar data distributions. However, when training data is limited, deep neural generative networks overfit and tend to replicate training samples. Prior works focus on few-shot image generation to produce high-quality and diverse results using a few target images. Unfortunately, abundant 3D shape data is typically hard to obtain as well. In this work, we make the first attempt to realize few-shot 3D shape generation by adapting generative models pre-trained on large source domains to target domains using limited data. To relieve overfitting and keep considerable diversity, we propose to maintain the probability distributions of the pairwise relative distances between adapted samples at feature-level and shape-level during domain adaptation. Our approach only needs the silhouettes of few-shot target samples as training data to learn target geometry distributions and achieve generated shapes with diverse topology and textures. Moreover, we introduce several metrics to evaluate the quality and diversity of few-shot 3D shape generation. The effectiveness of our approach is demonstrated qualitatively and quantitatively under a series of few-shot 3D shape adaptation setups.
\end{abstract}

\section{Introduction}
In recent years, 3D content has played significant roles in many applications, such as gaming, robotics, films, and animation. Currently, the most common method of creating 3D assets depends on manual efforts using specialized 3D modeling software like Blender \cite{blender} and Maya \cite{maya}, which is very time-consuming and cost-prohibitive to generate high-quality and diverse 3D shapes. As a result, the need for automatic 3D content generation becomes apparent.  

During the past decade, image generation has been widely studied and achieved great success using generative models, including generative adversarial networks (GANs) \cite{NIPS2014_5ca3e9b1, DBLP:conf/iclr/BrockDS19, Karras_2019_CVPR, Karras_2020_CVPR, Karras2021}, variational autoencoders (VAEs) \cite{kingma2013auto, rezende2014stochastic, vahdat2020nvae}, autoregressive models \cite{van2016conditional,chen2018pixelsnail,henighan2020scaling}, and diffusion models \cite{NEURIPS2020_4c5bcfec,song2020improved, dhariwal2021diffusion, nichol2021improved,kingma2021variational}. Compared with 2D images, 3D shapes are more complex and have different kinds of representations for geometry and textures. Inspired by the progress in 2D generative models, 3D generative models have become an active research area of computer vision and graphics and have achieved pleasing results in the generation of point clouds \cite{point2018,yang2019pointflow,zhou20213d}, implicit fields \cite{chen2019learning, mescheder2019occupancy}, textures \cite{pavllo2020convolutional,pavllo2021learning,richardson2023texture}, and shapes \cite{gao2022get3d, Liu2023MeshDiffusion}. In addition, recent works based on neural volume rendering \cite{mildenhall2021nerf} tackle 3D-aware novel view synthesis \cite{chan2021pi,chan2022efficient,gustylenerf,hao2021gancraft,niemeyer2021giraffe,or2022stylesdf,schwarz2020graf,xu20223d,zhou2021cips,schwarz2022voxgraf}. 

Similar to 2D image generative models like GANs and diffusion models, modern 3D generative models require large-scale datasets to avoid overfitting and achieve diverse results. Unfortunately, it is not always possible to obtain abundant data under some circumstances. Few-shot generation aims to produce diverse and high-quality generated samples using limited data. Modern few-shot image generation approaches \cite{wang2018transferring, ada,mo2020freeze, wang2020minegan, ewc, ojha2021few-shot-gan, zhao2022closer,zhu2022few,zhu2022ddpm, Zhao_2023_CVPR_fsig} adapt models pre-trained on large-scale source datasets to target domains using a few available training samples to relieve overfitting and produce adapted samples following target distributions. Nevertheless, few-shot 3D shape generation has yet to be studied, constrained by the complexity of 3D shape generation and the limited performance of early 3D shape generative models.

In this paper, we make the first attempt to study few-shot 3D shape generation pursuing high-quality and diverse generated shapes using limited data. We follow prior few-shot image generation approaches to adapt pre-trained source models to target domains using limited data. Since 3D shapes contain geometry and texture information, we need to clarify two questions: (i) what to learn from limited training data, and (ii) what to adapt from pre-trained source models to target domains. Naturally, we define two 3D shape domain adaptation setups: (i) geometry and texture adaptation (Setup A): the adapted models are trained to learn the geometry information of target data only and preserve the diversity of geometry and textures from source models, and (ii) geometry adaptation only (Setup B): the adapted models are trained to learn both the geometry and texture information of target data and preserve the diversity of geometry from source models only. Since the adaptation approach under setup A can be directly extended to setup B, we mainly focus on setup A and provide additional analysis and results of setup B in the supplementary.

We design a few-shot 3D shape generation approach based on modern 3D shape GANs, which synthesize textured meshes with randomly sampled noises requiring 2D supervision only. Source models directly fine-tuned on limited target data cannot maintain generation diversity and produce results similar to training samples. As shown in Fig. \ref{teaser}, two different source samples become analogous after few-shot domain adaptation, losing diversity of geometry and textures. Therefore, we introduce a pairwise relative distances preservation approach \cite{oord2018representation,ojha2021few-shot-gan,chen2020simple} to keep the probability distributions of geometry and texture pairwise similarities in generated shapes at both feature-level and shape-level during domain adaptation. In this way, the adapted models are guided to learn the common properties of limited training samples instead of replicating them. As a consequence, adapted models maintain similar generation diversity to source models and produce diverse results. 

The main contributions of our work are concluded as follows:
\begin{itemize}[leftmargin=0.4cm]
\item To our knowledge, we are the first to study few-shot 3D shape generation and achieve diverse generated shapes with arbitrary topology and textures.
\item We propose a novel few-shot 3D shape adaptation approach to learn target geometry distributions using 2D silhouettes of extremely limited data (e.g., 10 shapes) while preserving diverse information of geometry and textures learned from source domains.
\item We introduce several metrics to evaluate the quality and diversity of few-shot 3D shape generation and demonstrate the effectiveness of our approach qualitatively and quantitatively. 
\end{itemize}

\section{Related Work}
\subsection{3D Generative Models}
Early works \cite{wu2016learning,smith2017improved,lunz2020inverse,gadelha20173d,henzler2019escaping} extend 2D image generators to 3D voxel grids directly but fail to produce compelling results with high resolution due to the large computational complexity of 3D convolution networks. Other works explore the generation of alternative 3D shape representations, such as point clouds \cite{point2018,yang2019pointflow,zhou20213d} and implicit fields \cite{chen2019learning, mescheder2019occupancy}. Following works generate meshes with arbitrary topology using autoregressive models \cite{nash2020polygen} and GANs \cite{luo2021surfgen}. Meshdiffusion \cite{Liu2023MeshDiffusion} first applies diffusion models to generate 3D shapes unconditionally using 3D shapes for supervision. These works produce arbitrary topology only and need post-processing steps to achieve textured meshes which are compatible with modern graphics engines. 

DIBR \cite{chen2019learning} and Textured3DGAN \cite{pavllo2020convolutional,pavllo2021learning} synthesize textured 3D meshes based on input templated meshes, resulting in limited topology. GET3D \cite{gao2022get3d} first proposes a 3D generative model \cite{chang2015shapenet,turbosquid,renderpeople} to achieve arbitrary and diverse generation of 3D geometry structures and textures using 2D images for supervision. The proposed few-shot 3D shape generation approach is implemented with GET3D but is not confined to certain network architectures and can also be applied to other 3D shape generative models using 2D supervision.

\subsection{Few-shot Image Generation}
Few-shot image generation aims to produce high-quality images with great diversity utilizing only a few available training samples. Most modern approaches follow the TGAN \cite{wang2018transferring} method to adapt generative models pre-trained on large source domains, including ImageNet \cite{deng2009imagenet}, FFHQ \cite{Karras_2019_CVPR}, and LSUN \cite{yu2015lsun} et al., to target domains with limited data. Augmentation approaches \cite{tran2021data,zhao2020differentiable,zhao2020image} like ADA \cite{ada} help generate more different augmented training samples to relieve overfitting. BSA \cite{noguchi2019image} updates the scale and shift parameters in the generator and fixes the other parameters. FreezeD \cite{mo2020freeze} freezes the high-resolution layers in the discriminator to relieve overfitting. EWC \cite{ewc} applies elastic weight consolidation to regularize the generator by making it harder to change the critical weights which have higher Fisher information \cite{2017A} values. MineGAN \cite{wang2020minegan} adds additional networks to shift the distributions of the latent space of GANs by modifying the noise inputs of the generator. CDC \cite{ojha2021few-shot-gan} proposes a cross-domain consistency loss for generators and patch-level discrimination to build a correspondence between source and target domains. DCL \cite{zhao2022closer} uses contrastive learning to maximize the similarity between the corresponding source and target image pairs and push away the generated samples from training samples for greater diversity. MaskDis \cite{zhu2022few} proposes to regularize the discriminator using masked features and achieves outstanding visual effects. DDPM-PA \cite{zhu2022ddpm} first realizes few-shot image generation with diffusion models. 

Besides, other recent works have provided different research perspectives. RSSA \cite{xiao2022few} proposes a relaxed spatial structural alignment method using compressed latent space derived from inverted GANs \cite{Abdal_2020_CVPR}. AdAM \cite{adaptative} and RICK \cite{Zhao_2023_CVPR_fsig} achieve improvement in the adaptation of unrelated source/target domains. Research including MTG \cite{zhu2021mind}, OSCLIP \cite{kwon2022one}, GDA \cite{zhang2022generalized}, and DIFA \cite{zhang2022towards} et al. explore single-shot GAN adaptation with the guidance of pre-trained CLIP \cite{pmlr-v139-radford21a} image encoders. This work first explores few-shot 3D shape generation and shares similar ideas of preserving diverse information provided by source models, achieving the few-shot generation of diverse textured 3D shapes.

\section{Method}
\label{method}
Given 3D generative models pre-trained on large source domains, our approach adapts them to target domains by learning the common geometry properties of limited training data while maintaining the generation diversity of geometry and textures. Directly fine-tuned models tend to replicate training samples instead of producing diverse results since the deep generative networks are vulnerable to overfitting, especially when training data is limited. To this end, we propose to keep the probability distributions of the pairwise relative distances between adapted samples similar to source samples. 

\begin{figure}[t]
    \centering
    \includegraphics[width=1.0\linewidth]{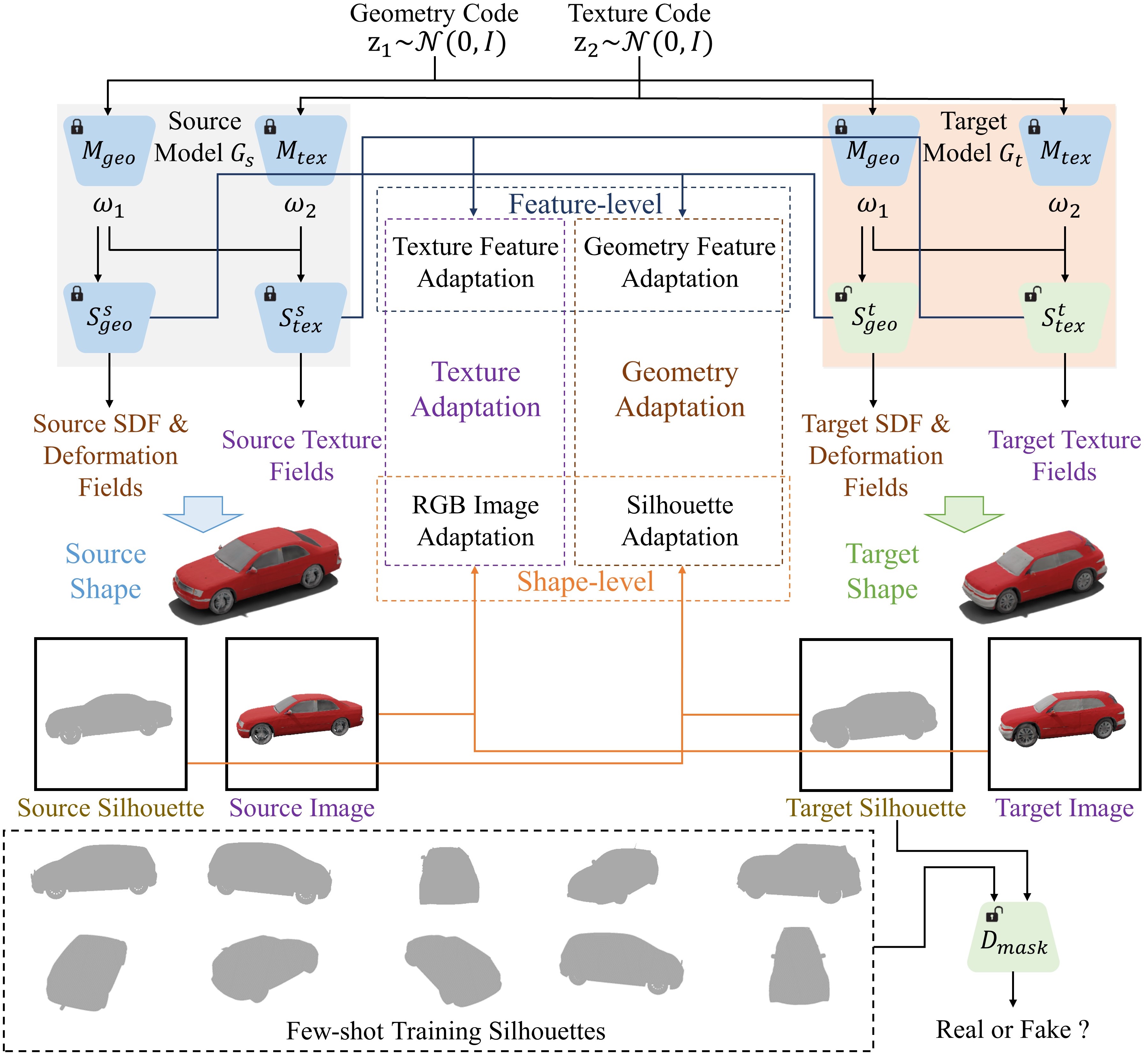}
    \caption{Overview of the proposed few-shot 3D shape generation approach using Cars $\rightarrow$ SUVs as an example: We maintain the distributions of pairwise relative distances between the geometry and textures of generated samples at feature-level and shape-level to keep diversity during domain adaptation. Only the silhouettes of few-shot target samples are needed as training data.}
    \label{overview}
\end{figure}

We employ the 3D shape generative model GET3D \cite{gao2022get3d} to illustrate the proposed approach, as shown in Fig. \ref{overview}.  GET3D realizes arbitrary generation of topology and textures using the combination of geometry and texture generators. Both generators are composed of mapping networks $M$ and synthesis networks $S$. GET3D utilizes the differentiable surface representation DMTet \cite{shen2021deep} to describe geometry with signed distance fields (SDF) defined on deformation fields \cite{gao2020beyond,gao2020learning}. The texture generator uses mapped geometry and texture codes as inputs and generates texture fields for explicit meshes obtained by adopting DMTet for surface extraction. GET3D is trained with two 2D discriminators applied to RGB images and silhouettes, respectively. Our approach can be divided into geometry adaptation (Sec. \ref{31}) and texture adaptation (Sec. \ref{32}) using source models as reference. Mapping networks of adapted models are fixed during domain adaptation. The silhouettes of target shapes are needed as training data to learn geometry distributions. Our approach is not tied to the network architectures of GET3D and is compatible with other 3D shape GANs using 2D supervision. 

\subsection{Geometry Adaptation}
\label{31}
We aim to guide adapted models to learn the common geometry properties of limited training samples while maintaining geometry diversity similar to source models. We propose to keep the probability distributions of pairwise relative distances between the geometry structures of adapted samples at feature-level and shape-level. We first sample a batch of geometry codes $\left\lbrace z_1^n \right\rbrace_{0}^{N}$ following the standard normal distribution $\mathcal{N}(0,I)$ and get mapped geometry latent codes $\left\lbrace \omega_1^n\right\rbrace^{N}_{0}$ using fixed geometry mapping networks $M_{geo}$. The probability distributions for the  $i^{th}$ noise vector $z_1^{i}$ in the source and target geometry generators at feature-level can be expressed as follows:
\begin{align}
    p_{geo,i}^{s,l} &= sfm(\left\lbrace sim(S^{s,l}_{geo}(\omega_1^i),S^{s,l}_{geo}(\omega_1^j))\right\rbrace_{\forall i\neq j}), \\
    p_{geo,i}^{t,l} &= sfm(\left\lbrace sim(S^{t,l}_{geo}(\omega_1^i),S^{t,l}_{geo}(\omega_1^j))\right\rbrace_{\forall i\neq j}),
\end{align}
where $sfm$ and $sim$ represent the softmax function and cosine similarity between activations at the $l^{th}$ layer of the source and target geometry synthesis networks which generate SDF and deformation fields. Then we guide target geometry synthesis networks to keep similar probability distributions to source models during domain adaptation with the feature-level geometry loss:
\begin{align}
    \mathcal{L}_{geo}(S^{s}_{geo}, S^{t}_{geo})=\mathbb{E}_{z_1^{i}\sim \mathcal{N}(0,I)} \sum_{l,i} D_{KL}(p_{geo,i}^{t,l}||p_{geo,i}^{s,l}),
\end{align}
where $D_{KL}$ represents KL-divergence. Similarly, we use source and target silhouettes in place of the features in geometry synthesis networks to keep the pairwise relative distances of adapted samples at shape-level. For this purpose, we further sample a batch of texture codes $\left\lbrace z_2^n \right\rbrace_{0}^{N}$ for shape generation. The probability distributions of shapes generated from the $i^{th}$ noise vectors ($z_1^{i}$ and $z_2^{i}$) by the source and target generators are given by:
\begin{align}
    p_{mask,i}^{s} &= sfm(\left\lbrace sim(Mask(G_s(z_1^{i},z_2^{i})), Mask(G_s(z_1^{j},z_2^{j}))) \right\rbrace_{\forall i\neq j}), \\
    p_{mask,i}^{t} &= sfm(\left\lbrace sim(Mask(G_t(z_1^{i},z_2^{i})), Mask(G_t(z_1^{j},z_2^{j}))) \right\rbrace_{\forall i\neq j}), 
\end{align}
where $G_s$ and $G_t$ are the source and target shape generators, $Mask$ represents the masks of 2D rendered shapes. We have the shape-level mask loss for geometry adaptation as follows:
\begin{align}
    \mathcal{L}_{mask}(G_s,G_t)=\mathbb{E}_{z_1^{i}, z_2^{i}\sim \mathcal{N}(0,I)} \sum_{i}D_{KL}(p_{mask,i}^{t}||p_{mask,i}^{s}).
\end{align}

\subsection{Texture Adaptation}
\label{32}
In addition, we also encourage adapted models to preserve the texture information learned from source domains and generate target shapes with diverse textures. We still apply the pairwise relative distances preservation approach to relieve overfitting and keep the generation diversity of textures. Since the generated textures for explicit meshes contain both geometry and texture information, we propose to use textures in regions shared by two generated shapes to compute the pairwise relative distances of textures while alleviating the influence of geometry. In the same way, we use the randomly sampled geometry codes $\left\lbrace z_1^n \right\rbrace_{0}^{N}$ and texture codes $\left\lbrace z_2^n \right\rbrace_{0}^{N}$ and get mapped latent codes $\left\lbrace \omega_1^n \right\rbrace_{0}^{N}$ and $\left\lbrace \omega_2^n \right\rbrace_{0}^{N}$ with fixed geometry and texture mapping networks $M_{geo}$ and $M_{tex}$, respectively. The shared regions of two generated shapes produced by the source and adapted models are defined as the intersection of the masks of the 2D rendered shapes:
\begin{align}
     M^{s}_{i,j} &= Mask(G_s(z_1^{i},z_2^{i})) \land Mask(G_s(z_1^{j},z_2^{j})) \; (i\neq j), \\
     M^{t}_{i,j} &= Mask(G_t(z_1^{i},z_2^{i})) \land Mask(G_t(z_1^{j},z_2^{j})) \; (i\neq j).
\end{align}
The probability distributions for the  $i^{th}$ noise vectors ($z_1^{i}$ and $z_2^{i}$) in the source and target texture generators at feature-level can be expressed as follows:
\begin{align}
    p_{tex,i}^{s,m} &= sfm(\left\lbrace sim(S_{tex}^{s,m}(\omega_1^{i},\omega_2^{i})\otimes M^{s}_{i,j} ,S_{tex}^{s,m}(\omega_1^{j},\omega_2^{j})\otimes M^{s}_{i,j}) \right\rbrace_{\forall i\neq j}), \\
    p_{tex,i}^{t,m} &= sfm(\left\lbrace sim(S_{tex}^{t,m}(\omega_1^{i},\omega_2^{i}) \otimes M^{t}_{i,j},S_{tex}^{t,m}(\omega_1^{j},\omega_2^{j}) \otimes M^{t}_{i,j}) \right\rbrace_{\forall i\neq j}),
\end{align}
where $\otimes$ and $sim$ represent the element-wise multiplication of tensors and cosine similarity between activations at the $m^{th}$ layer of the source and target texture synthesis networks. For shape-level texture adaptation, we use 2D rendered shapes of RGB formats in place of the features in texture synthesis networks to compute the probability distributions:
\begin{align}
    p_{rgb,i}^{s} &= sfm(\left\lbrace sim(RGB(G_s(z_1^i,z_2^i))\otimes M^{s}_{i,j}, RGB(G_s(z_1^j,z_2^j))\otimes M^{s}_{i,j}) \right\rbrace_{\forall i\neq j}), \\
    p_{rgb,i}^{t} &= sfm(\left\lbrace sim(RGB(G_t(z_1^i,z_2^i))\otimes M^{t}_{i,j}, RGB(G_t(z_1^j,z_2^j))\otimes M^{t}_{i,j}) \right\rbrace_{\forall i\neq j}),
\end{align}
where $RGB$ represents the rendered RGB images of generated shapes. We have the feature-level texture loss and shape-level RGB loss for texture adaptation as follows:
\begin{align}
    \mathcal{L}_{tex}(S_{tex}^{s},S_{tex}^{t}) &= \mathbb{E}_{z_1^{i}, z_2^{i}\sim \mathcal{N}(0,I)} \sum_{m,i}D_{KL} (p_{tex,i}^{t,m}||p_{tex,i}^{s,m}), \\
    \mathcal{L}_{rgb}(G_s,G_t) &= \mathbb{E}_{z_1^{i}, z_2^{i}\sim \mathcal{N}(0,I)} \sum_{i}D_{KL}(p_{rgb,i}^{t}||p_{rgb,i}^{s}).
\end{align}

\subsection{Overall Optimization Target}
Since adapted models are guided to learn the geometry information of training data, we only use the mask discriminator and apply the above-mentioned pairwise relative distances preservation methods to preserve diverse geometry and texture information learned from source domains. In this way, our approach only needs the silhouettes of few-shot target shapes as training data. The overall optimization target $\mathcal{L}$ of adapted models is defined as follows:
\begin{equation}
    \begin{aligned}
        \mathcal{L} =  \mathcal{L}(D_{mask},G_t)+\mu \mathcal{L}_{reg} & + \mu_1 \mathcal{L}_{geo}(S^{s}_{geo}, S^{t}_{geo}) + \mu_2 \mathcal{L}_{mask}(G_s,G_t) \\ & + \mu_3  \mathcal{L}_{tex}(S_{tex}^{s},S_{tex}^{t}) + \mu_4 \mathcal{L}_{rgb}(G_s,G_t).
    \end{aligned}
    \label{loss_all}
\end{equation}
Here $\mathcal{L}(D_{mask},G_t)$ and $\mathcal{L}_{reg}$ represent the adversarial objective of silhouettes and regularization term of generated SDFs used in GET3D. More details of these two losses are added in Appendix \ref{get3dloss}. $\mu, \mu_1, \mu_2, \mu_3, \mu_4$ are hyperparameters set manually to control the regularization levels.   

\section{Experiments}
\label{experiments}
We employ a series of few-shot 3D shape adaptation setups to demonstrate the effectiveness of our approach. We first show the qualitative results in Sec. \ref{41}. Then we introduce several metrics to evaluate quality and diversity quantitatively in Sec. \ref{42}. Finally, we ablate our approach in Sec. \ref{43}.

\textbf{Basic Setups} Our approach is evaluated with GET3D \cite{gao2022get3d}. The hyperparameter of SDF regularization $\mu$ is set as 0.01 for all experiments. We empirically find 
$\mu_1 = 2e+4, \mu_2=5e+3, \mu_3=5e+3, \mu_4=1e+4$ to work well for the employed adaptation setups. We conduct experiments with batch size 4 on a single NVIDIA A40 GPU. The learning rates of the generator and discriminator are set as 0.0005. The adapted models are trained for 40K-60K iterations. The resolution of 2D rendered RGB images and silhouettes is 1024$\times$1024. More details of implementation are added in Appendix \ref{appendix_implement}. 

 
\textbf{Datasets} We use ShapeNetCore Cars and Chairs \cite{chang2015shapenet} as source datasets and sample several 10-shot shapes as target datasets, including (i) Trucks, (ii) Racing Cars, (iii) Sport Utility Vehicles (SUVs), (iv) Police Cars, (v) Ambulances corresponding to Cars and (vi) Rocking Chairs, (vii) Modern Chairs, (viii) Lawn Chairs corresponding to Chairs. Police Cars and Ambulances are used for the experiments of geometry adaptation (see Appendix \ref{setup2}). Other datasets are applied to the experiments of geometry and texture adaptation. The training data are rendered using 24 randomly sampled and evenly distributed camera poses. All the few-shot target datasets are visualized in Appendix \ref{datasets}.


\textbf{Baselines} Since few existing works explore few-shot 3D shape generation, we compare the proposed approach with directly fine-tuned methods (DFTM) and fine-tuned models using fixed texture generators (FreezeT), including fixed texture mapping and texture synthesis networks.

\begin{figure}[t]
    \centering
    \includegraphics[width=1.0\linewidth]{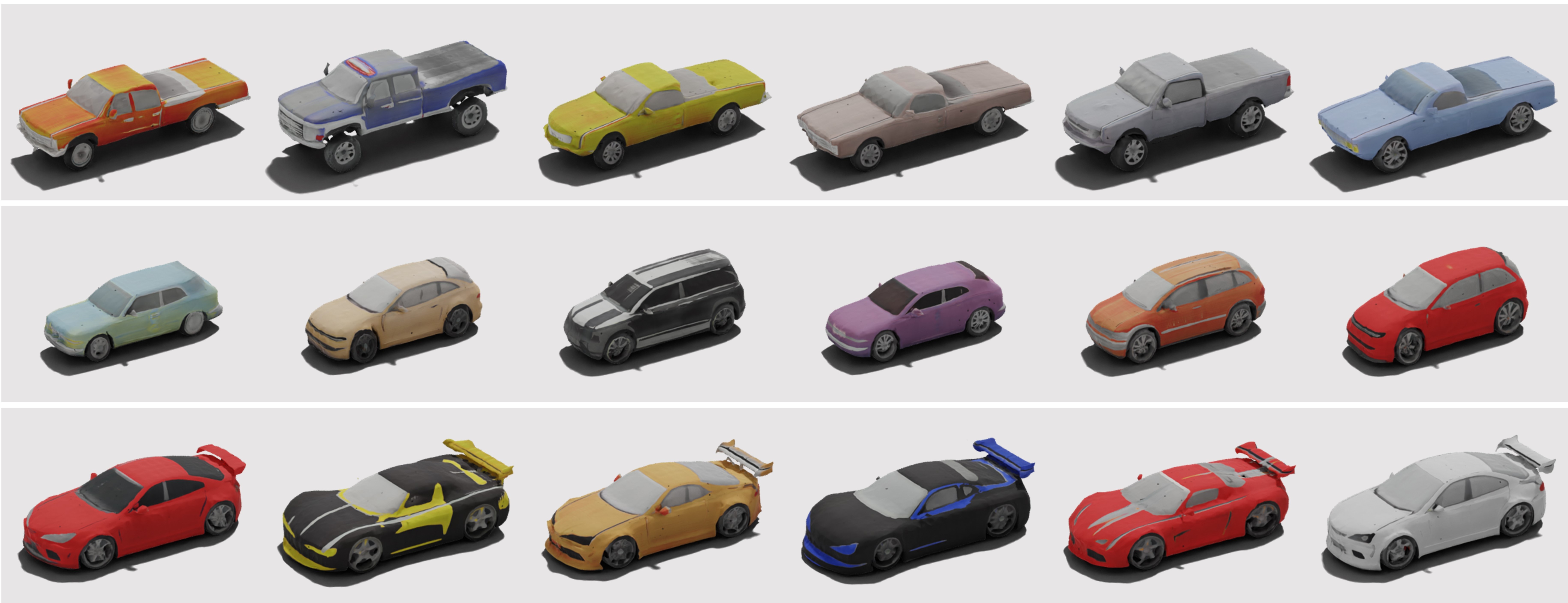}
    \caption{10-shot generated shapes of our approach on Cars $\rightarrow$ Trucks, SUVs, and Racing Cars.}
    \label{cars}
\end{figure}

\begin{figure}[t]
    \centering
    \includegraphics[width=1.0\linewidth]{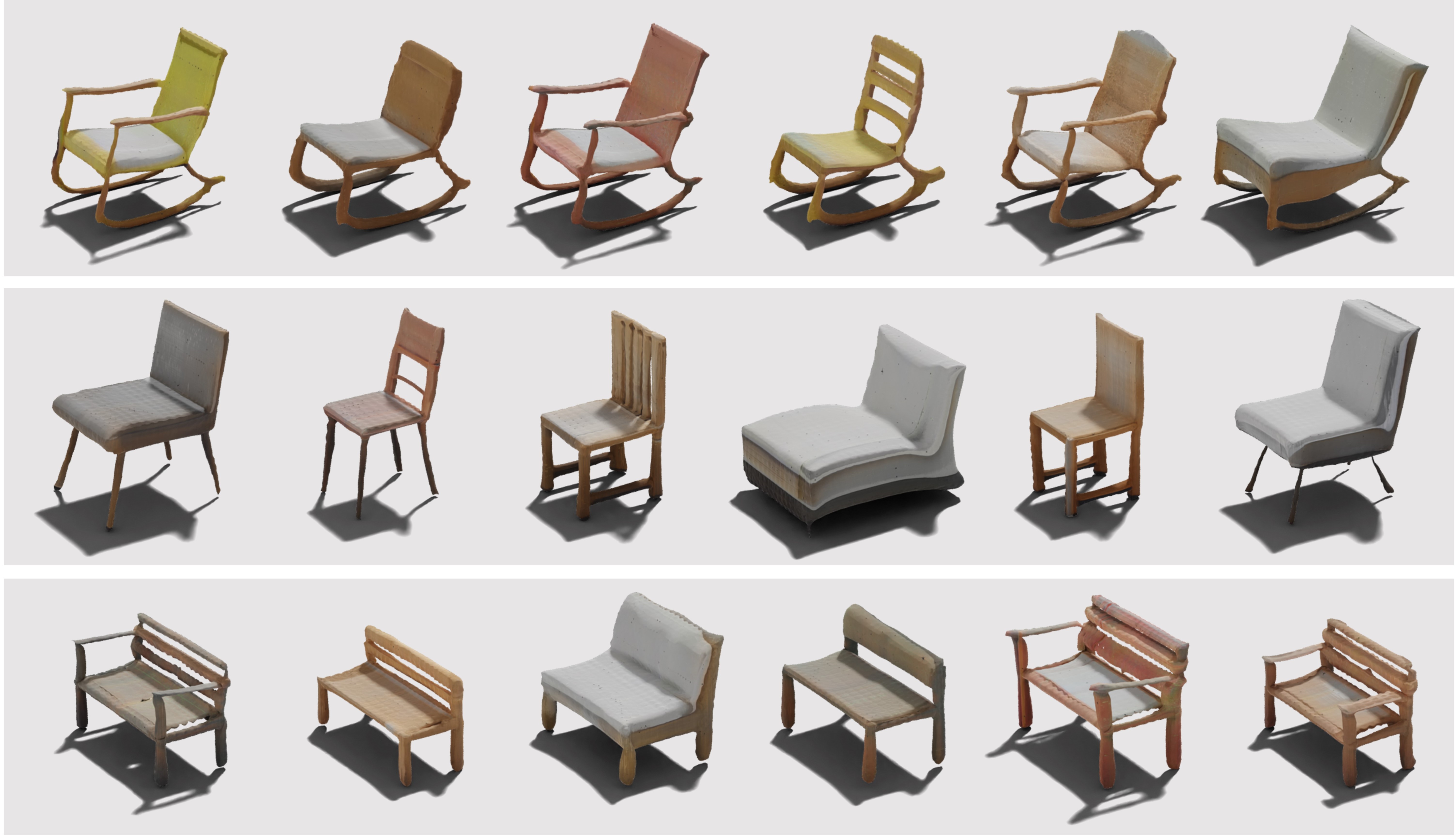}
    \caption{10-shot generated shapes of our approach on Chairs $\rightarrow$ Rocking Chairs, Modern Chairs, and Lawn Chairs.}
    \label{chairs}
\end{figure}

\begin{figure}[t]
    \centering
    \includegraphics[width=1.0\linewidth]{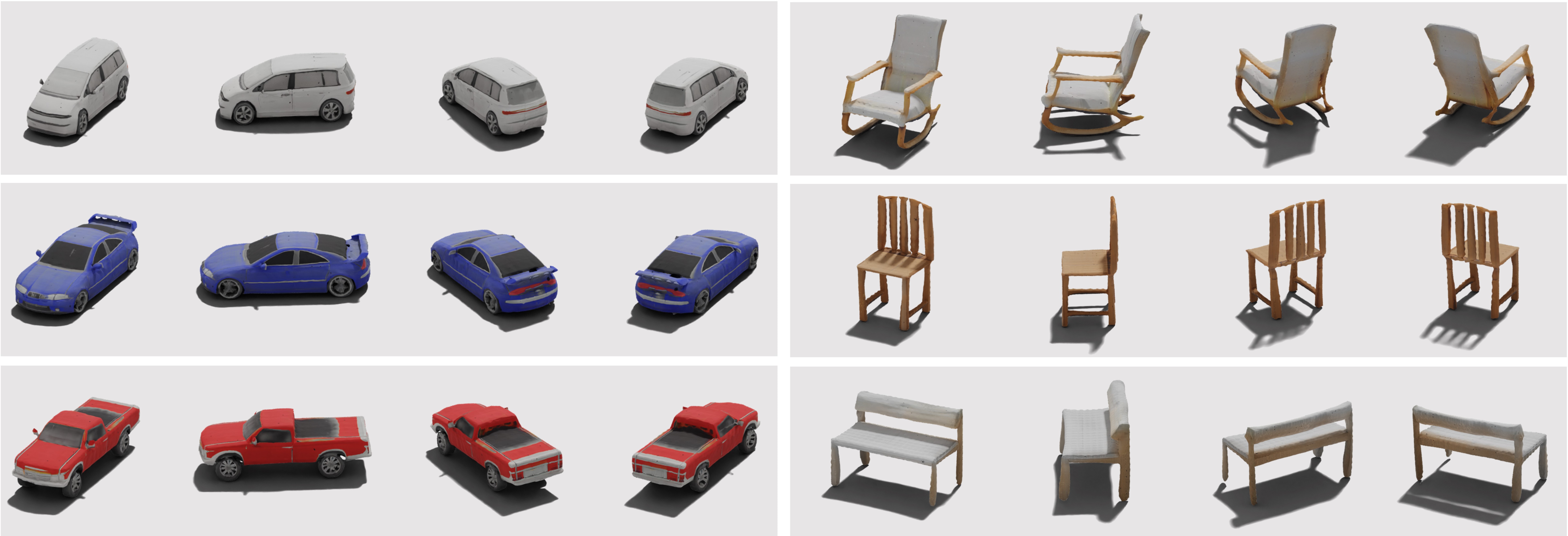}
    \caption{Multi-view rendered shapes produced by our approach on different 10-shot target domains.}
    \label{multiview}
\end{figure}

\begin{figure}[t]
    \centering
    \includegraphics[width=1.0\linewidth]{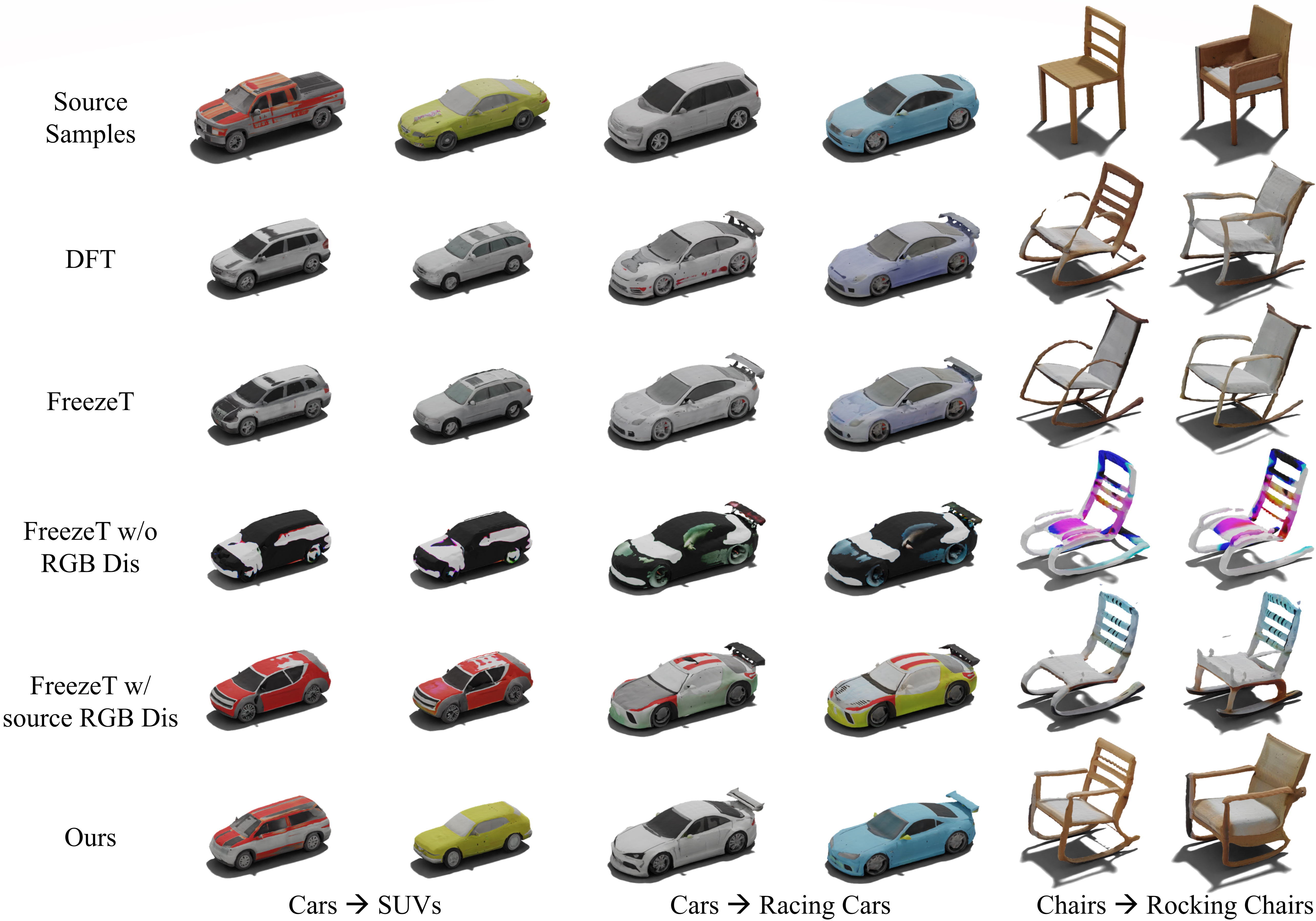}
    \caption{Visualization samples comparison on 10-shot Cars $\rightarrow$ SUVs, Cars $\rightarrow$ Racing Cars, and Chairs $\rightarrow$ Rocking Chairs. Results of different approaches are synthesized with fixed noise inputs.}
    \label{compare}
\end{figure}

\subsection{Qualitative Evaluation}
\label{41}
We visualize the samples produced by our approach using source models pre-trained on ShapeNetCore Cars and Chairs in Fig. \ref{cars} and \ref{chairs}, respectively. Our approach only needs the silhouettes of few-shot training samples as target datasets to adapt source models to target domains while maintaining generation diversity of geometry and textures. In Fig. \ref{multiview}, we add generated shapes of different target domains rendered in multiple views. Our approach produces high-quality results different from the few-shot training samples. In addition, we compare the proposed approach with baselines using fixed noise inputs for intuitive comparison in Fig. \ref{compare}. DFTM models replicate training samples and fail to keep generation diversity. FreezeT also fails to produce diverse textures since the mapped geometry codes influence the fixed texture synthesis networks. As a result, FreezeT models produce textured meshes similar to training samples under the guidance of RGB discriminators. Therefore, we further train FreezeT models without RGB discriminators or using source RGB discriminators. However, these two approaches still fail to preserve the diverse geometry and texture information of source models and cannot produce reasonable shapes. Our approach maintains the pairwise relative distances between generated shapes at feature-level and shape-level. It achieves high-quality and diverse adapted samples sharing geometry and texture information with source samples.

\begin{table}[t]
\centering
\caption{Quantitative evaluation of our approach. Generated shapes of different approaches are synthesized from fixed noise inputs for fair comparison. CD scores are multiplied by $10^3$. The best results are highlighted in bold. Our approach performs better on both generation quality and diversity.}
\small
\begin{tabular}{c|c|c|c|c|c|c}
\hline
Datasets & Approach & CD ($\downarrow$) & Intra-CD ($\uparrow$) & Pairwise-CD ($\uparrow$) & Intra-LPIPS ($\uparrow$) & Pairwise-LPIPS ($\uparrow$)\\
\hline
\multirow{3}*{\shortstack{Cars $\rightarrow$ \\SUVs}} & DFTM & $1.401$ & $0.316 \pm 0.002$ & $0.513 \pm 0.001$ & $0.062 \pm 0.001$ & $0.063 \pm 0.012$ \\
 & FreezeT & $1.553$ & $0.240 \pm 0.005$ & $0.326 \pm 0.002$ & $0.055 \pm 0.002$ & $0.060 \pm 0.014$ \\
 & Ours & $\pmb{1.323}$ & $\pmb{0.511 \pm 0.006}$ & $\pmb{0.814 \pm 0.007}$ & $\pmb{0.109 \pm 0.026}$ & $\pmb{0.095 \pm 0.022}$ \\
\hline
\multirow{3}*{\shortstack{Cars $\rightarrow$ \\Trucks}} & DFTM & $4.014$ & $0.441 \pm 0.003$ & $0.689 \pm 0.003$ & $0.112 \pm 0.002$ & $0.119 \pm 0.024$ \\
 & FreezeT & $4.175$ & $0.412 \pm 0.006$ & $0.766 \pm 0.002$ & $0.120 \pm 0.003$ & $0.128 \pm 0.027$ \\
 & Ours & $\pmb{3.940}$ & $\pmb{1.061 \pm 0.014}$ & $\pmb{1.175 \pm 0.004}$ & $\pmb{0.145 \pm 0.022}$ & $\pmb{0.146 \pm 0.033}$ \\
\hline
\multirow{3}*{\shortstack{Chairs $\rightarrow$ \\ Lawn \\ Chairs}} & DFTM & $40.559$ & $4.001 \pm 0.005$ & $13.598 \pm 0.013$ & $0.165 \pm 0.029$ & $0.141 \pm 0.047$\\
 & FreezeT & $39.422$ & $4.671 \pm 0.022$ & $19.269 \pm 0.024$ & $0.120 \pm 0.032$ & $0.165 \pm 0.040$ \\
 & Ours & $\pmb{38.661}$ & $\pmb{5.852 \pm 0.031}$ & $\pmb{22.989 \pm 0.022}$ & $\pmb{0.278 \pm 0.040}$ & $\pmb{0.166 \pm 0.054}$ \\
\hline
\multirow{3}*{\shortstack{Chairs $\rightarrow$ \\Rocking \\ Chairs}} & DFTM & $18.996$ & $7.405 \pm 0.022$ & $15.312 \pm 0.011$ & $0.202 \pm 0.039$ & $0.203 \pm 0.037$ \\
 & FreezeT & $18.503$ & $5.541 \pm 0.014$ & $11.977 \pm 0.009$ & $0.203 \pm 0.046$ & $0.204 \pm 0.036$ \\
 & Ours & $\pmb{17.598}$ & $\pmb{8.773 \pm 0.029}$ & $\pmb{16.165 \pm 0.015}$ & $\pmb{0.289 \pm 0.062}$ & $\pmb{0.222 \pm 0.063}$ \\
\hline
\end{tabular}
\label{fid}
\end{table}

\subsection{Quantitative Evaluation}
\label{42}
\textbf{Evaluation Metrics} The generation quality of adapted models represents their capability to learn target geometry distributions. Chamfer distance (CD) \cite{chen2003visual} is employed to compute the distances of geometry distributions between 5000 adapted samples and target datasets containing relatively abundant target data to obtain reliable results. Besides, we design several metrics based on CD and LPIPS \cite{zhang2018unreasonable} to evaluate the diversity of geometry and textures in adapted samples, respectively. LPIPS measures the perceptual distances between images. Evaluation metrics for generation diversity are computed in two ways: (i) pairwise-distances: we randomly generate 1000 shapes and compute the pairwise distances averaged over them, (ii) intra-distances \cite{ojha2021few-shot-gan}: we first assign the generated shapes to one of the few-shot training samples with the lowest LPIPS distance and then compute the average pairwise distances within each cluster averaged over all the clusters. LPIPS results are averaged over 8 evenly distributed views of rendered shapes. Adapted models which tend to replicate training samples may achieve fine pairwise distances but only get intra-distances close to 0. Adapted models with great generation diversity achieve large values of both pairwise and intra-distances.  

The quantitative results of our approach are compared with baselines under several few-shot adaptation setups, as listed in Table \ref{fid}. Our approach learns target geometry distributions better in terms of CD. Moreover, our approach also performs better on all the benchmarks of generation diversity, indicating its strong capability to produce diverse shapes with different geometry structures and textures. 

\begin{figure}[t]
    \centering
    \includegraphics[width=1.0\linewidth]{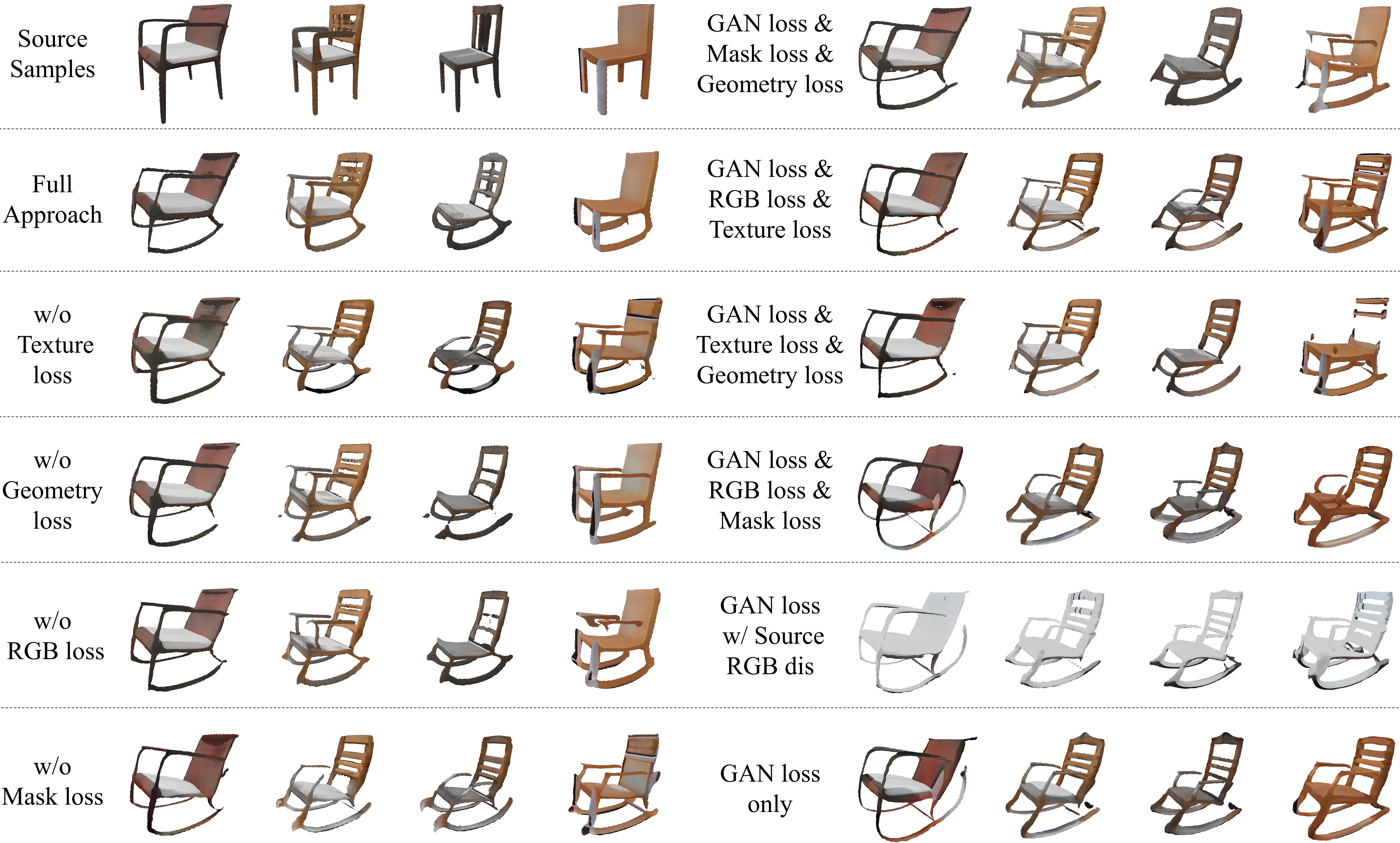}
    \caption{Qualitative ablations of our approach using 10-shot Chairs $\rightarrow$ Rocking Chairs as an example. Results of different approaches are synthesized with fixed noise inputs for intuitive comparison.}
    \label{ablation}
\end{figure}

\subsection{Ablation Analysis}
\label{43}
Our approach is composed of the pairwise relative distances preservation methods applied to geometry and textures at feature-level and shape-level. We provide ablation analysis to show the roles played by each component of our approach. In Fig. \ref{ablation}, we show the qualitative ablation analysis using 10-shot Chairs $\rightarrow$ Rocking Chairs as an example. Our full approach adapts source samples to target domains while preserving diverse geometry and texture information. Adapted models only using GAN loss with mask discrimination fail to maintain geometry diversity or produce high-quality shapes. Adding fixed source RGB discriminators results in texture degradation. Absence of the feature-level texture loss makes adapted models harder to preserve the texture information learned from source domains. Absence of shape-level RGB loss leads to repetitive textures and discontinuous shapes. As for the feature-level geometry and shape-level mask losses, their absence results in adapted samples sharing similar geometry structures and incomplete shapes. We also add ablations using geometry and mask losses, texture and RGB losses, feature-level losses, and shape-level losses, respectively. None of these approaches generate compelling results with diverse topology and textures. Incomplete geometry structures and low-quality textures can be found in their adapted samples. Moreover, the full approach also achieves quantitative results better than other settings, as shown in Appendix \ref{suppablations}. 

\section{Conclusion and Limitations}
This paper first explores few-shot 3D shape generation. We introduce a novel domain adaptation approach to produce 3D shapes with diverse topology and textures. The relative distances between generated samples are maintained at both feature-level and shape-level. We only need the silhouettes of few-shot target samples as training data to learn target geometry distributions while keeping diversity. Our approach is implemented based on GET3D to demonstrate its effectiveness. However, it is not constrained by specific network architectures and can be combined with more powerful 3D shape generative models using 2D supervision to produce higher-quality results in the future. Despite the compelling results of our approach, it still has some limitations. Firstly, it sometimes cannot completely preserve the diverse textures of source samples. Besides, it is mainly designed for related source/target domains. Extending our approach to unrelated domain adaptation would be promising. Nevertheless, we believe this work takes a further step towards democratizing 3D content creation by transferring knowledge in available source models to fit target distributions using few-shot data.  





\small
\bibliographystyle{abbrv}
\bibliography{refs}
\normalsize

\clearpage



\appendix

\section{Broader Impact}
We propose a novel approach for few-shot 3D shape generation, achieving diverse 3D shape generation using limited training data. Our approach is more prone to biases introduced by training data than typical artificial intelligence generative models since it only needs silhouettes of few-shot samples to train adapted models. The proposed approach is applicable to 3D shape generative models and not tailored for sensitive applications like generating human bodies. Therefore, we recommend practitioners to apply abundant caution when dealing with such applications to avoid problems of races, skin tones, or gender identities.

\begin{figure}[t]
    \centering
    \includegraphics[width=1.0\linewidth]{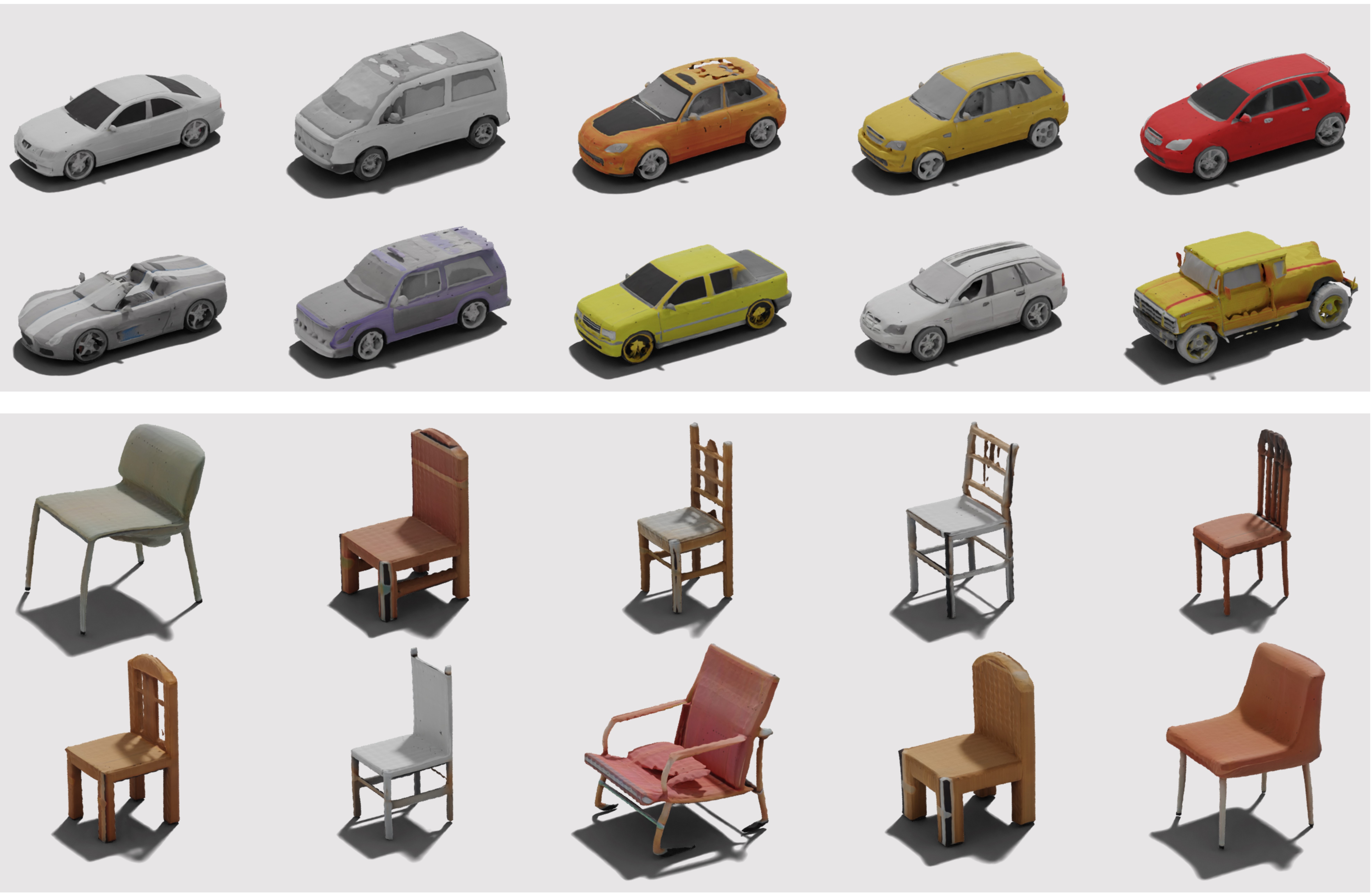}
    \caption{Generated shapes produced by the source GET3D models trained on ShapeNetCore Cars and Chairs datasets.}
    \label{source}
\end{figure}


\section{More Details of GET3D}
\label{get3dloss}
GET3D \cite{gao2022get3d} is the first 3D shape generative model to produce textured meshes with arbitrary topology and textures. Here we add more details of the GET3D model. The mapping networks of GET3D are composed of 3D convolutional and fully connected networks. The synthesis networks for SDF and deformation fields are MLPs. As for the texture synthesis networks, GET3D uses generator network structures similar to StyleGAN2 to generate textures using triplane feature maps as inputs. GET3D also follows StyleGAN2 to use the same 2D discriminators and non-saturating GAN objective. Two 2D image discriminators are applied to RGB images and silhouettes, respectively. Given x representing an RGB image or a silhouette, the adversarial objective is defined as:
\begin{align}
    \mathcal{L}(D_x,G_t) = \mathbb{E}_{z \in \mathcal{N}}[g(D_x(R(G_t(z))))]+\mathbb{E}_{I_x\in p_x}\left[g(-D_x(I_x))+\lambda||\nabla D_x(I_x)||^2_2 \right],
\end{align}
where $g(u)=-log(1+exp(-u))$, $p_x$ and $R$ represent the real image distributions and rendering functions for RGB images or silhouettes. In Eq. \ref{loss_all}, we employ the discriminator for silhouettes as $\mathcal{L}(D_{mask},G_t)$. The discriminator for RGB images used in GET3D is expressed as $\mathcal{L}(D_{rgb}, G_t)$. The regularization loss $\mathcal{L}_{reg}$ in Eq. \ref{loss_all} is designed to remove internal floating surfaces since GET3D aims to generate textured meshes without internal structures. $\mathcal{L}_{reg}$ is defined as a cross-entropy loss between the SDF values of neighboring vertices \cite{munkberg2022extracting}: 
\begin{align}
    \mathcal{L}_{reg}=\sum_{i,j\in \mathbb{S}_e, i\neq j}H(\sigma(s_i),sign(s_j))+H(\sigma(s_j),sign(s_i)).
\end{align}
Here $H$ and $\sigma$ represent binary cross-entropy loss and sigmoid function. $s_i, s_j$ are SDF values of neighboring vertices in the set of unique edges $\mathbb{S}_{e}$ in the tetrahedral grid. The regularization loss $\mathcal{L}_{reg}$ is applied to all the experiments (including ablation analysis) in this paper.

\begin{figure}[t]
    \centering
    \includegraphics[width=1.0\linewidth]{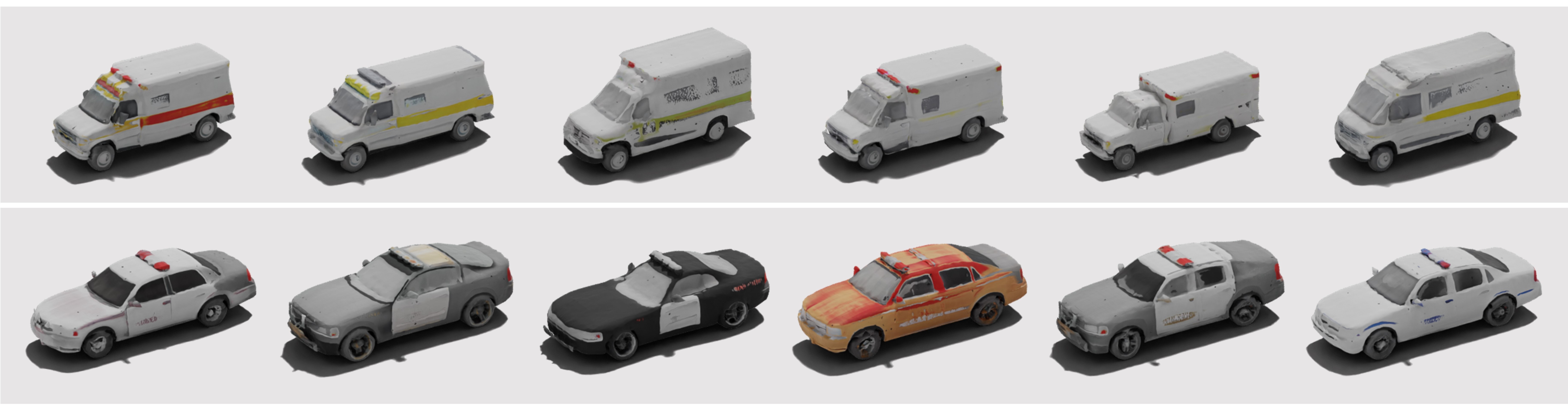}
    \caption{10-shot generated shapes of our approach on Cars $\rightarrow$ Ambulances and Police Cars.}
    \label{results3}
\end{figure}

\begin{table}[t]
\centering
\caption{Quantitative evaluation of our approach on generation quality of geometry and textures.}
\begin{tabular}{c|c|c|c}
\hline
Datastes & Approach & FID ($\downarrow$) & CD ($\downarrow$) \\
\hline
\multirow{2}*{\shortstack{Cars $\rightarrow$ \\Ambulances}} & DFTM & $\pmb{101.583}$ & $6.896$ \\
& Ours & $103.708$ & $\pmb{5.963}$ \\
\hline
\multirow{2}*{\shortstack{Cars $\rightarrow$ \\ Police Cars}} & DFTM & $86.833$ & $6.440$ \\
& Ours & $\pmb{74.958}$ & $\pmb{5.616}$ \\
\hline
\end{tabular}
\label{fid3}
\end{table}

GET3D needs multi-view rendered RGB images and silhouettes with corresponding camera distribution parameters as training data. Therefore, it is evaluated with synthetic datasets such as ShapeNetCore \cite{chang2015shapenet} and TurboSquid \cite{turbosquid}. Future work may extend GET3D to single-view real-world datasets. If so, our approach can be applied to the advanced models to realize few-shot generation of real-world 3D shapes using single-view silhouettes.


In Fig. \ref{source}, we provide generated samples of the officially released GET3D models trained on ShapNetCore Cars and Chairs datasets. These models are used as source models in our experiments. GET3D generates shapes with arbitrary topology and textures. However, improvement room still exists for better results, such as incomplete textures of tires. As a result, our approach produces some samples with incomplete textures of tires, as shown in Fig. \ref{cars}. Our approach can be combined with better generative models in the future to achieve better visual effects.

\begin{table}[t]
\centering
\caption{Quantitative evaluation of our approach on generation diversity of geometry and textures.}
\small
\begin{tabular}{c|c|c|c|c|c}
\hline
Datastes & Approach & Intra-CD ($\uparrow$) & Pairwise-CD ($\uparrow$) & Inra-LPIPS ($\uparrow$) & Pairwise-LPIPS ($\uparrow$)\\
\hline
\multirow{2}*{\shortstack{Cars $\rightarrow $ \\ Ambulances}} & DFTM & $0.300\pm 0.002$ & $\pmb{1.027 \pm 0.007}$ & $0.079 \pm 0.009$ & $0.083 \pm 0.017$ \\
& Ours & $\pmb{0.558 \pm 0.004}$ & $0.638\pm 0.006$ & $\pmb{0.093 \pm 0.018}$ & $\pmb{0.086 \pm 0.016}$  \\
\hline
\multirow{2}*{\shortstack{Cars $\rightarrow$ \\ Police Cars}} & DFTM & $0.426 \pm 0.003$ & $\pmb{0.926 \pm 0.008}$ & $0.109 \pm 0.002$ & $0.108 \pm 0.017$ \\
 & Ours & $\pmb{0.902 \pm 0.005}$ & $0.902 \pm 0.006$ & $\pmb{0.115 \pm 0.009}$ & $\pmb{0.120 \pm 0.020}$ \\
 \hline
\end{tabular}
\label{fid4}
\end{table}

\begin{table}[t]
\centering
\caption{Quantitative ablations of the proposed approach using 10-shot Chairs $\rightarrow$ Rocking Chairs as an example. The full approach performs the best on both generation quality and diversity.}
\setlength\tabcolsep{5.0pt}
\small
\begin{tabular}{c|c|c|c|c|c}
\hline
Approach & CD ($\downarrow$) & Intra-CD ($\uparrow$) & Pairwise-CD ($\uparrow$) & Intra-LPIPS ($\uparrow$) & Pairwise-LPIPS ($\uparrow$)\\
\hline
w/o Texture loss & $18.178$ & $8.054 \pm 0.028$ & $13.533 \pm 0.010$ & $0.221 \pm 0.013$ & $0.210 \pm 0.045$ \\
w/o Geometry loss & $18.409$ & $7.551 \pm 0.019$ & $12.549 \pm 0.009$ & $0.271 \pm 0.023$ & $0.217 \pm 0.057$ \\
w/o RGB loss & $17.762$ & $7.207 \pm 0.018$ & $13.124 \pm 0.010$ & $0.211 \pm 0.006$ & $0.213 \pm 0.034$ \\
w/o Mask loss & $18.275$ & $6.878 \pm 0.014$ & $12.435 \pm 0.008$ & $0.248 \pm 0.010$ & $0.208 \pm 0.010$ \\
Full Approach & $\pmb{17.598}$ & $\pmb{8.773 \pm 0.029}$ & $\pmb{16.165 \pm 0.015}$ & $\pmb{0.289 \pm 0.062}$ & $\pmb{0.222 \pm 0.063}$ \\
\hline
\end{tabular}
\label{fid2}
\end{table}

\section{Geometry Adaptation Only}
\label{setup2}
In this section, we add the discussion of geometry adaptation only (Setup B). Source models are trained to learn geometry and textures from limited training data under setup B. Adapted models preserve the diversity of geometry learned from source domains. As for textures, we guide adapted models to fit the distributions of training samples. 

\textbf{Method} The proposed adaptation approach under setup B has two differences compared with setup A (texture and geometry adaptation) discussed in our paper. Firstly, the feature-level texture loss and shape-level RGB loss are no longer needed. Secondly, generators are guided by the RGB discriminator to learn target texture distributions. Therefore, we need RGB images of rendered real samples as inputs for the RGB discriminator. The overall optimization target of adapted models under setup B is defined as follows:
\begin{align}
    \begin{aligned}
        \mathcal{L} = \mathcal{L}(D_{mask},G_t)+\mathcal{L}(D_{rgb},G_t)+\mu \mathcal{L}_{reg} +  \mu_1 \mathcal{L}_{geo}(S^{s}_{geo}, S^{t}_{geo}) + \mu_2 \mathcal{L}_{mask}(G_s,G_t)
    \end{aligned}
\end{align}
We follow GET3D to set $\mu = 0.01$ and empirically find $\mu_1$ and $\mu_2$ ranging from 2e+3 to 1e+4 appropriate for the adaptation setups used in our paper. 

\begin{figure}[t]
    \centering
    \includegraphics[width=1.0\linewidth]{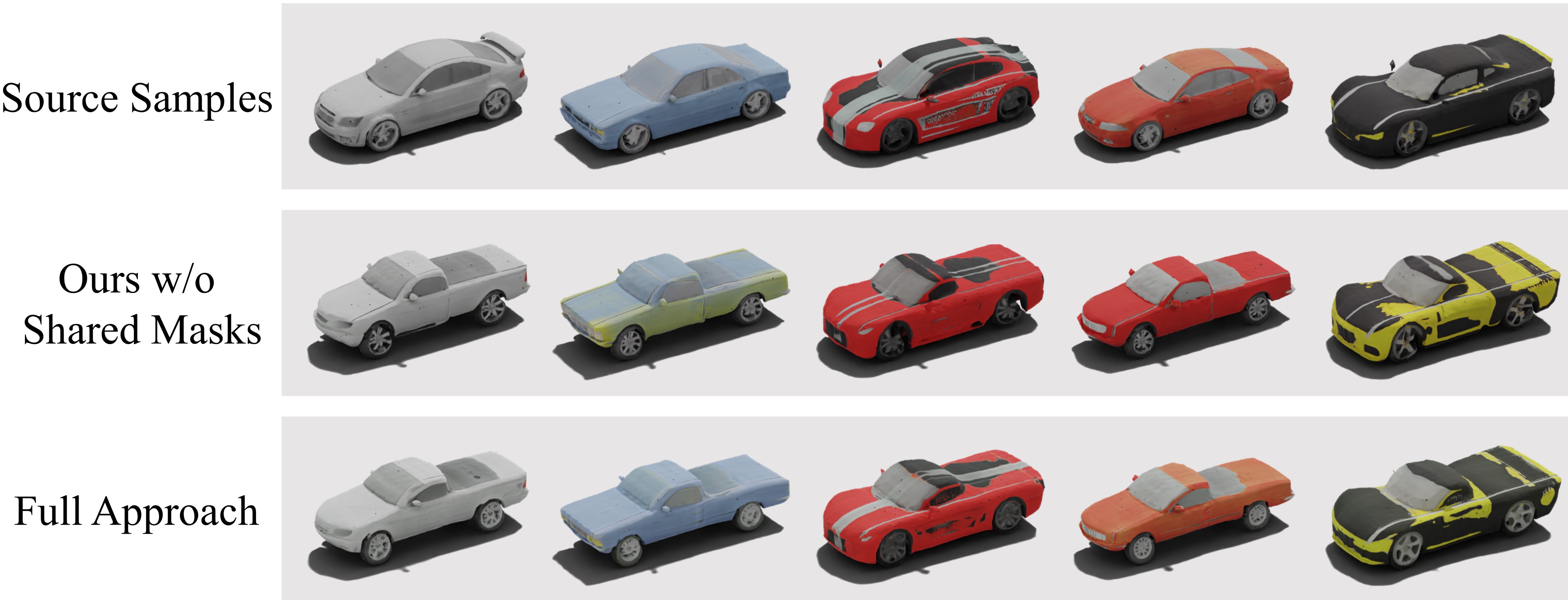}
    \caption{Qualitative ablations of shared masks applied to the feature-level texture loss and shape-level RGB loss using 10-shot Cars $\rightarrow$ Trucks as an example. The generated shapes of different approaches are synthesized with fixed noise inputs for intuitive comparison.}
    \label{ablation4}
\end{figure}

\textbf{Experiments} We sample two 10-shot target datasets from ShapeNetCore \cite{chang2015shapenet} to evaluate our approach under setup B, including Ambulances and Police Cars in correspondence to the source domain Cars. The basic setups of experiments under setup B are consistent with those under setup A (see Sec. \ref{experiments}). We provide qualitative and quantitative results of our approach to demonstrate its effectiveness under setup B. As shown in Fig. \ref{results3}, our approach produces ambulances and police cars with diverse topology using few-shot training samples qualitatively. For quantitative evaluation, we further add FID \cite{heusel2017gans} to evaluate the generation quality. FID results are averaged over 24 views of rendered shapes. The quantitative results are listed in Tables \ref{fid3} and \ref{fid4}. Compared with DFTM models, our approach performs better on learning target geometry distributions in terms of CD. As for FID, our approach achieves better results on Cars $\rightarrow$ Police Cars and gets results close to the DFTM model on Cars $\rightarrow$ Ambulances. Besides, our approach achieves greater generation diversity in terms of Intra-CD and Intra-LPIPS. DFTM models get better results on Pairwise-CD and results close to our approach on Pairwise-LPIPS but get apparently worse results on intra-distances, indicating that they are overfitting to few-shot training samples and tend to replicate them instead of producing diverse results. We do not include FreezeT models for comparison under setup B since the adapted models are trained to learn the texture information from limited training samples.

\begin{figure}[t]
    \centering
    \includegraphics[width=1.0\linewidth]{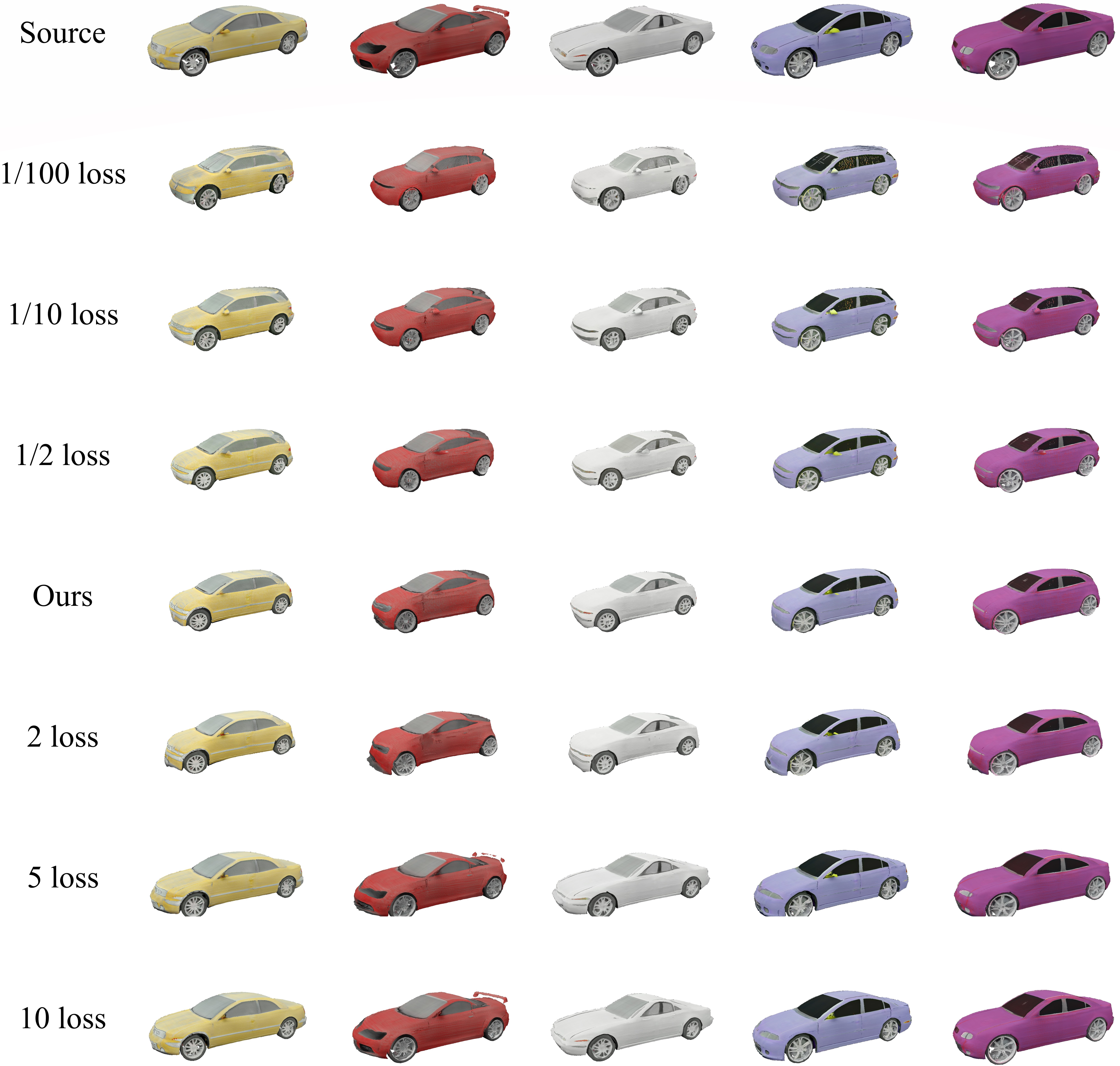}
    \caption{Qualitative ablations of the hyperparameters applied to the proposed adaptation losses using 10-shot Cars $\rightarrow$ SUVs as an example.}
    \label{ablation3}
\end{figure}

\begin{figure}[t]
    \centering
    \includegraphics[width=1.0\linewidth]{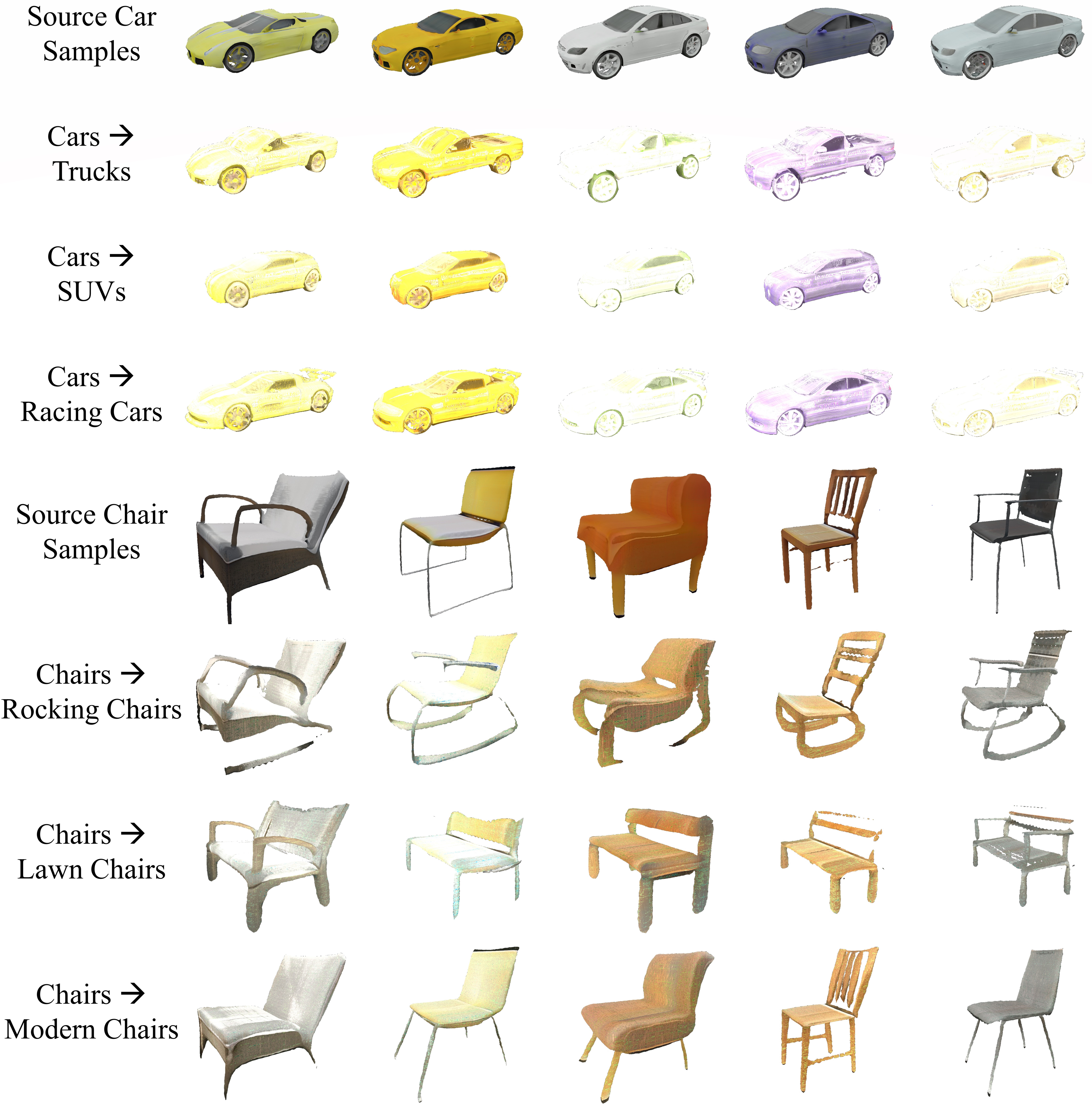}
    \caption{Ablations of fixed mapping networks during domain adaptation. Without fixed mapping networks, our approach fails to preserve the diverse texture information of source samples and produces blurred textures.}
    \label{ablation5}
\end{figure}

\begin{figure}[t]
    \centering
    \includegraphics[width=1.0\linewidth]{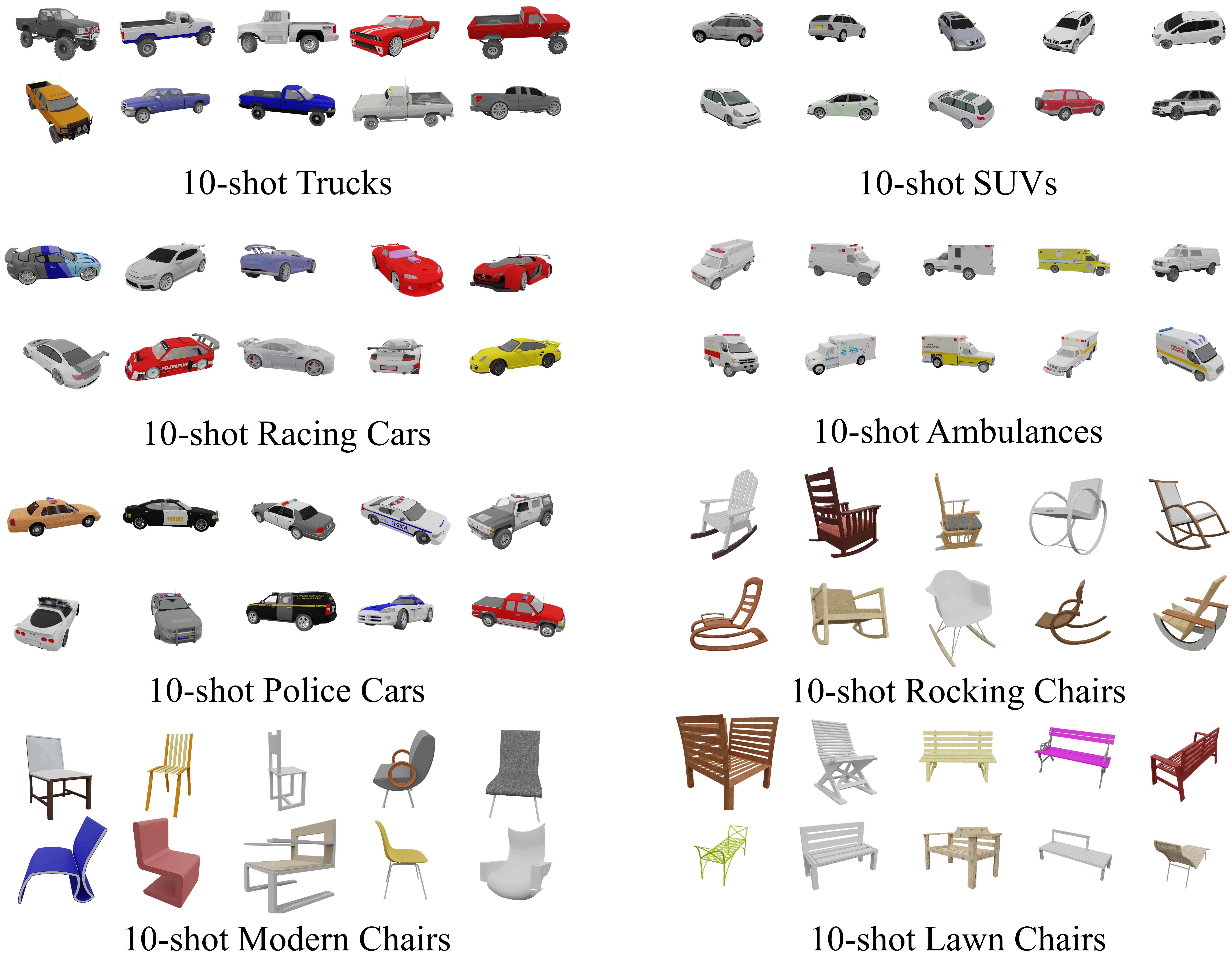}
    \caption{Visualization of the 10-shot 3D shape datasets used in this paper. Using randomly sampled views, we provide one 2D rendered RGB image for each training shape.}
    \label{data}
\end{figure}

\section{Supplementary Ablations}
\label{suppablations}
\textbf{Quantitative Ablations} Table \ref{fid2} shows quantitative ablations of our approach. The full approach achieves the best quantitative results on both generation quality and diversity. Without feature-level geometry loss or shape-level mask loss, adapted models performs worse on geometry diversity in terms of Intra-CD and Pairwise-CD. Similarly, adapted models perform worse on texture diversity in terms of Intra-LPIPS and Pairwise-LPIPS without feature-level texture loss or shape-level RGB loss. 

\textbf{Ablations of Shared Masks} In addition, we provide qualitative ablations for the shared masks used for feature-level texture loss and shape-level RGB loss computation in Fig. \ref{ablation4}. Absence of shared masks causes geometry structures to bias the domain adaptation of textures, making the textures of adapted samples more different from source samples. For example, the blue and orange source cars change into yellow-blue and red trucks during the 10-shot domain adaptation. The full approach applies shared masks to relieve the influence of geometry structures and achieves better preservation of the texture information in source models.

\textbf{Ablations of Hyperparameters} We add ablations of the hyperparameters applied to the proposed four adaptation losses. We use different values of hyperparameters and provide qualitative results using 10-shot Cars $\rightarrow$ SUVs in Fig. \ref{ablation3}. Too large values of hyperparameters prevent adapted models from learning target distributions, resulting in results similar to source samples. Too small values of hyperparameters lead to diversity degradation of geometry and textures. We empirically recommend hyperparameters $\mu_1,\mu_2,\mu_3,\mu_4$ ranging from 2e+3 to 1e+4 for adaptation setups used in this paper.

\textbf{Ablations of Fixed Mapping Networks} As illustrated in Sec. \ref{method}, the geometry and texture mapping networks $M_{geo}$ and $M_{tex}$ are fixed during domain adaptation. We propose this design to isolate the geometry and texture adaptation since the texture synthesis networks need the mapped geometry codes as inputs. Without fixed mapping networks, fine-tuned geometry mapping networks would influence the texture adaptation process. We add ablations of fixed mapping networks under different adaptation setups and provide qualitative samples in Fig. \ref{ablation5}. The low-quality adapted samples show blurred textures and fail to preserve the diverse texture information of source samples.

\begin{table}[t]
\centering
\caption{The time cost of our approach trained for 1K iterations in terms of seconds on a single NVIDIA A40 GPU (image resolution $1024\times 1024$, batch size 4).}
\begin{tabular}{c|l|c}
\hline
Setups & Approaches & \makecell[c]{Time cost for 1K iterations}  \\
\hline
\multirow{7}*{Setup A} & DFTM & 228.83\\
& GAN loss only & 272.27\\
& GAN loss w/ Texture loss & 352.80 \\
& GAN loss w/ Geometry loss & 295.28\\
& GAN loss w/ RGB loss & 291.62\\
& GAN loss w/ Mask loss & 279.34\\
& Full Approach & 392.67\\
\hline
\multirow{5}*{Setup B} & DFTM & 281.15\\
& GAN loss only & 316.55 \\
& GAN loss w/ Geometry loss & 344.82\\
& GAN loss w/ Mask loss & 322.51 \\
& Full Approach & 340.38\\
\hline
\end{tabular}
\label{time}
\end{table}

\section{More Details of Datasets}
\label{datasets}
This paper employs several 10-shot datasets sampled from ShapeNetCore \cite{chang2015shapenet} as training data for few-shot 3D shape generation. The rendered datasets of randomly sampled views are shown in Fig. \ref{data}. As for the main experiments of our paper, we only need silhouettes of target samples as training data, as shown in Fig. \ref{overview}. For the experiments of geometry adaptation only (Sec. \ref{setup2}), rendered RGB images are also needed to train adapted models. 

We employ CD \cite{chen2003visual} and FID \cite{heusel2017gans} as quantitative evaluation metrics for generation quality. Datasets containing relatively abundant data are applied for evaluation to obtain reliable results. The few-shot samples are excluded from the relatively abundant datasets to avoid the influence of overfitting. The relatively abundant Trucks, SUVs, Ambulances, Police Cars, Rocking Chairs, and Lawn Chairs datasets contain 40, 369, 73, 133, 87, and 78 samples. 

\section{Computational Cost}
\label{time_section}
Table \ref{time} shows the computational cost of our approach under two adaptation setups using a single NVIDIA A40 GPU. We also ablate our approach to show the computational cost of each component. The adapted models are trained for about 40K-60K iterations in our experiments, costing about 4.4-6.5 and 3.8-5.7 hours under setup A (geometry and texture adaptation) and setup B (geometry adaptation only), respectively. DFTM under setup B is the same as training GET3D models directly. DFTM under setup A excludes the RGB discriminator. Compared with DFTM, the approach only using GAN loss includes the time cost by source models.  

\section{Inspiration of Loss Design}
Our approach is composed of feature-level geometry loss, feature-level texture loss, shape-level mask loss, and shape-level RGB loss sharing similar formats to preserve the relative distances between generated shapes. Our approach is mainly inspired by contrastive learning methods \cite{oord2018representation, he2020momentum, chen2020simple}. Similar approaches can be found in recent few-shot image generation approaches \cite{ojha2021few-shot-gan,zhu2022few,zhu2022ddpm} as well. This paper first explores few-shot 3D shape generation and proposes an effective domain adaptation approach by adopting the pairwise relative distances preservation loss for geometry and textures at feature-level and shape-level.

\section{More Details of Implementation}
\label{appendix_implement}
The proposed approach is implemented based on the official code of GET3D \cite{gao2022get3d}. The setups of adapted models are consistent with those of the officially released source models trained on ShapeNetCore Cars and Chairs \cite{chang2015shapenet}. The geometry and texture synthesis networks are composed of 2-layers MLP networks. We concatenate the output features of the first layers in the synthesis networks of SDFs and deformation fields for feature-level geometry loss computation since the output features of the second layers have different sizes for SDFs and deformation fields. We also use the features in the synthesis networks of SDFs and deformation fields separately for feature-level geometry loss computation. Unfortunately, it is more time-consuming and fails to produce better results. For feature-level texture loss computation, we use the output features of the second layers in the texture synthesis network, which has the same resolution as the generated shapes. Therefore, we can directly apply the shared masks of generated shapes to the texture features.

The weights in target models are initialized to source models. We set the learning rates of the generator and discriminator as 0.0005, which is lower than the learning rates of source models (0.002), to realize more refined adaptation processes. We set the hyperparameters of the proposed losses ($\mu_1,\mu_2,\mu_3,\mu_4$) equally for adaptation from Cars and Chairs and achieve high-quality results. Different hyperparameters can be tried to obtain compelling results under other adaptation setups. We train adapted models with batch size 4 on a single NVIDIA A40 GPU (45GB GPU memory). Our approach needs about 20 GB GPU memory for the image resolution of $1024\times 1024$. The standard deviations of pairwise-distance and intra-distance results listed in Tables \ref{fid}, \ref{fid2}, and \ref{fid4} are computed across shape pairs picked from generated samples and 10 clusters (the same number as few-shot training samples), respectively.

\begin{figure}[t]
    \centering
    \includegraphics[width=1.0\linewidth]{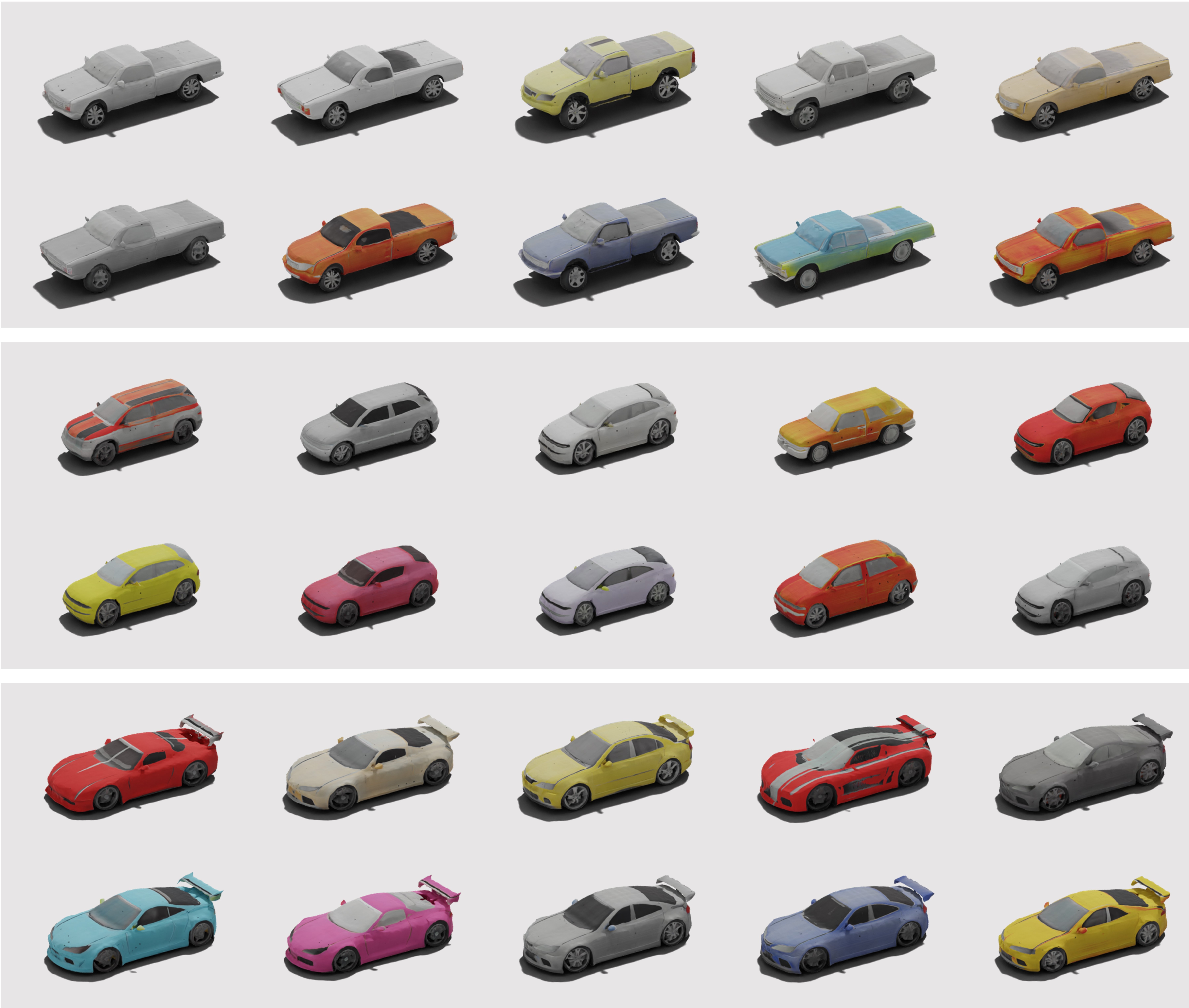}
    \caption{Additional 10-shot generated shapes of our approach on Cars $\rightarrow$ Trucks, SUVs, and Racing Cars. }
    \label{results4}
\end{figure}

\begin{figure}[t]
    \centering
    \includegraphics[width=1.0\linewidth]{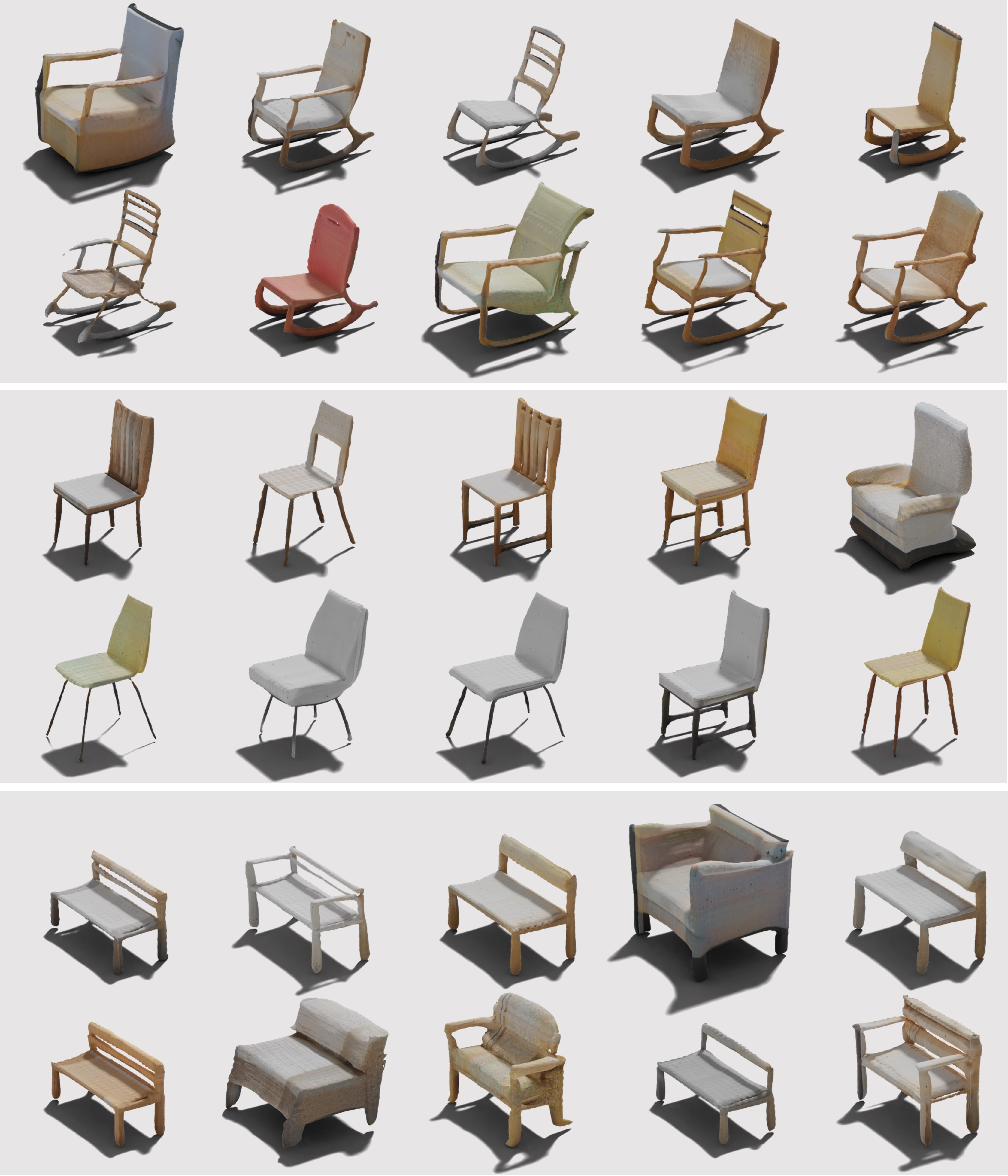}
    \caption{Additional 10-shot generated shapes of our approach on Chairs $\rightarrow$ Rocking Chairs, Modern Chairs, and Lawn Chairs.}
    \label{results5}
\end{figure}

\section{More Visualization Results}
\label{appendix_examples}
As supplements to generated samples shown in Fig. \ref{cars} and \ref{chairs}, we show more examples produced by our approach under several few-shot adaptation setups. Adapted samples obtained with the source models pre-trained on ShapeNetCore Cars and Chairs \cite{chang2015shapenet} are shown in Fig. \ref{results4} and \ref{results5}, respectively.


\end{document}